\pdfoutput=1

\documentclass[11pt]{article}

\usepackage{acl}

\usepackage{times}
\usepackage{latexsym}

\usepackage[T1]{fontenc}

\usepackage[utf8]{inputenc}

\usepackage{microtype}

\usepackage{booktabs}
\usepackage{multirow}

\usepackage{inconsolata}
\usepackage{graphicx}
\usepackage{amsmath} 
\usepackage{amssymb}
\usepackage{booktabs}
 
\usepackage{textcomp}
\usepackage{mathtools}
\usepackage{graphicx}  
\usepackage{arydshln}

\usepackage{mathrsfs}
\usepackage{multirow}
\usepackage{subcaption}
\usepackage{xcolor}
\usepackage{soul}

\usepackage{tcolorbox}
\usepackage{caption}
\usepackage[inline]{enumitem}

\newcommand{\todo}[1]{\textcolor{red}{TODO: #1}} 

\newenvironment{dialogue}{%
    \par\medskip
    \noindent\ignorespaces
}{%
    \par\medskip
}

\newcommand{\speak}[1]{\textbf{#1}: }
\title{Okay, Let's Do This! Modeling Event Coreference with Generated Rationales and Knowledge Distillation}

\author{Abhijnan Nath, Shadi Manafi, Avyakta Chelle\and Nikhil Krishnaswamy \\
Situated Grounding and Natural Language (SIGNAL) Lab\\
        Department of Computer Science \\ Colorado State University, Fort Collins, Colorado, USA \\ \{abhijnan.nath, nkrishna\}@colostate.edu
        }

\begin{document}
\maketitle
\begin{abstract}
In NLP, Event Coreference Resolution (ECR) is the task of connecting event clusters that refer to the same underlying real-life event, usually via neural systems. In this work, we investigate using abductive free-text rationales (FTRs) generated by modern autoregressive LLMs as distant supervision of smaller student models for cross-document coreference (CDCR) of events. We implement novel rationale-oriented event clustering and knowledge distillation methods for event coreference scoring that leverage enriched information from the FTRs for improved CDCR without additional annotation or expensive document clustering.  Our model using coreference-specific knowledge distillation achieves SOTA $B^3$ $F_1$ on the ECB+ and GVC corpora and we establish a new baseline on the AIDA Phase 1 corpus. Our code can be found at \url {https://github.com/csu-signal/llama_cdcr}.

\end{abstract}

\section{Introduction}

Event Coreference Resolution (ECR) is the task of connecting mentions that refer to the same underlying real-life event. Since descriptions of similar events often use similar words in similar context, systems can achieve strong baseline performance simply by comparing the lemmas of the event triggers~\cite{bugert2021crossdocument,nath2023axomiyaberta}. 

However, most ECR datasets contain many event pairs that might be coreferent despite different lemmas, or non-coreferent despite similar lemmas or tokens. Consider this example: ``{\it Video of Brooklyn woman's fatal shooting is \underline{played} at trial of two men charged in rooftop \underline{gunplay.}}'' The lexical similarity in the event triggers is misleading since they are actually \textit{not} coreferent. Typical heuristic-based systems fail in such cases if the decision is made at the \textit{event pair level}.

Intuitively, a human might solve a challenging coreference problem by engaging in a step-by-step "inner monologue" that reasons about the context, participants, actions, locations, etc., in both events~\cite{bershon1992cooperative,alderson2015inner}. Recent works like InstructGPT ~\cite{ouyang2022training} have shown that generative large language models (LLMs) can engage in Chain-of-Thought reasoning, a kind of step-by-step reasoning that appears human-like. Such models have also demonstrated abductive reasoning capabilities about relations between events \cite{zhao2023abductive, ravi2023happens} and zero-shot resolution abilities in various coreference benchmarks like CoNLL-2012 and ECB+~\cite{yang-etal-2022-gpt, le2023large}.

In this paper, we seek to model such an inner-monologue or a step-by-step reasoning process in an event coreference system. We augment existing coreference corpora using a novel instruction-based zero-shot prompting framework~\cite{kojima2023large} that guides a generative LLM to produce outputs displaying abductive reasoning about the coreference samples therein.  



These intermediate reasoning steps, consolidated into free-text rationales (FTRs) for the coreference labels of the mention pairs, are then used to guide a two-stage modeling procedure. We first perform "Rationale-Oriented Event Clustering (ROEC)" by directly optimizing a "student" model to encode cluster-level information in the coreference graph. Since the generated rationales comes from a disparate model distribution, we simultaneously align event pairs with their corresponding FTRs in the student model's latent space. We then use the optimized student distribution as our backbone encoder, which learns coreference probabilities of event pairs by jointly optimizing the task-supervision component with additional supervision from the generative LLM "teacher" distribution using the rationales as soft labels.
Our novel contributions are:
  \begin{itemize}
    \item{A method for augmenting coreference datasets with evidence for decisions using FTRs from state-of-the-art generative LLMs;}
    \item{A novel clustering method to align these rationales with corresponding event pairs;}
    \item{A novel optimization framework for distilling contextual cues from these rationales into smaller encoder models using a customized loss function.}

 
 \end{itemize}

Rationales provide an additional soft supervisory signal through which we train the student model with additional information about contextual cues for event coreference, but are not required at inference, increasing our approach's generalizability. We evaluate our method on three event CDCR corpora: Event Coreference Bank Plus (ECB+), the Gun Violence Corpus (GVC), and AIDA Phase 1. Our method achieves state-of-the-art $B^3$ score on ECB+ and GVC, and sets a novel performance baseline on AIDA Phase 1, without a document clustering step as used in many other methods. We perform detailed ablations of individual components of our method and evaluate how each one contributes to the final performance. Our code, weights, and FTR sets will be released upon publication.

\section{Related Work}
\label{sec:related}

\paragraph{Event CDCR}


Most previous works in CDCR address the challenge of pairing across documents with a document clustering step that reduces the search space for potential candidates and ensures tractability of pairwise computations~\cite{lee2012joint, yang2015hierarchical, choubey2017event,cattan2021cross, caciularu2021cdlm, yu2022pairwise}. However, preprocessing with document clustering misses a non-trivial amount of coreferring pairs between clusters~\cite{cremisini-finlayson-2020-new}. \citet{bugert2020breaking} also demonstrate that this tends to overfit CDCR systems to corpora like ECB+~\cite{cybulska2014using}, that have unrealistic lexical distinction between topics, thus reducing systems' generalizability to corpora with more referential ambiguity (e.g., GVC;~\citet{vossen2018don}). \citet{held-etal-2021-focus} avoid document clustering by modeling discourse focus at the mention level, and \citet{ahmed20232} leverage heuristics that capture discourse level lemmatic features.


While \citet{held-etal-2021-focus} also suggest using knowledge distillation techniques~\cite{gou2021knowledge} to enhance pairwise computations even further, using a heuristic allows us to approach knowledge distillation using a single encoder to model both cluster-level and pairwise signals without using separate encoders for candidate retrieval and candidate scoring.  In line with this, we use \citet{ahmed20232}'s heuristic-based approach for candidate generation during training and inference.


\paragraph{Abductive Reasoning and Free-Text Rationales}

Previous work on abductive reasoning in coreference resolution has used rule-based and entity-specific approaches, with causal relations between events only appearing as supporting evidence for the coreference decision~\cite{raina2005robust,inoue2012coreference,yamamoto2015boosting}. \citet{zhao2023abductive} and \citet{bhagavatula2019abductive} explore Bayesian methods to model the plausibility of abductive explanations and suggest mutual exclusivity of such explanations. The rise of autoregressive general-purpose LLMs have inspired works like \citet{ho2022large, snell2022learning, shridhar2023distilling, narang2020wt5,sun2022investigating} and \citet{rajani2019explain}, which use LLM-generated FTRs for NLI tasks like commonsense reasoning, for both training and inference. \citet{wiegreffe2021measuring} and \citet{west-etal-2022-symbolic} explore knowledge distillation techniques using such FTRs and suggest evaluation frameworks to assess their quality. \citet{ahmed-etal-2024-xtool,ahmed-etal-2024-xamr} explores capabilities of LLMs like GPT to generate argumental and temporal information between entities in order to augment the annotation process in event corefernece corpora like the ECB+.

Research in cognitive psychology~\cite{alderson2015inner} suggests that a well-developed "inner monologue" is crucial for different aspects of problem-solving, often as a “working memory” or a cognitive rehearsal tool~\cite{sokolov2012inner}. \citet{ravi2023happens} use this strategy for temporal reasoning in GPT-3-generated FTRs to improve coreference computations. However, due to the black-box nature of LLMs like GPT-3, their method assumes the FTRs themselves as the complete knowledge, with no access to the LLM's internal distribution. In contrast, our research uses an open-weight LLM both to generate step-by-step FTRs with reasoning about coreference, and so that we can access the underlying model distribution.

\section{Method}
{
\begin{figure*}[t]
  \includegraphics[width=\textwidth]{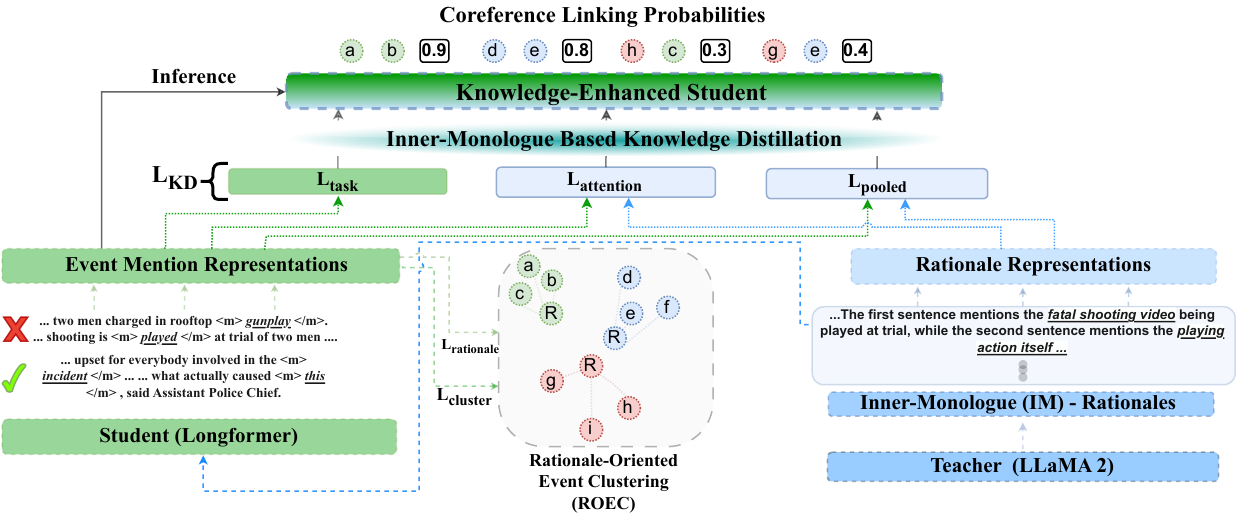}
  \vspace*{-5mm}
\caption{Schematic system overview: Step-by-step FTRs resembling an "inner-monologue" are generated using an LLM (teacher model) conditioned on the gold coreference label. FTRs are then clustered along with event pairs to optimize the student's latent space (ROEC). The optimized student learns further coreference-specific contextual cues in the rationales from the teacher's latent space. Arrows show the gradient flow during training from the teacher (blue) and student (green) during the ROEC (dashed) and knowledge distillation (dotted) phases, respectively. Solid black line indicates inference samples, which include no rationale text or signal from the teacher model. Letters $a$-$i$ in the ROEC block represent distinct event mentions, and the colors represent an event cluster (such that all the blue circles cluster together). “R” represents a set of rationales that justify the linking of different mentions in a single cluster. }  \label{fig:full_pipeline}
\end{figure*}}
 
Our method for event coreference resolution consists of three parts:
\begin{enumerate*}[label=\arabic*)]
    \item Using curated zero-shot instructions specifying the ECR task, we generate abductive rationales from a generative LLM teacher model;
    \item We implement a Rationale-Oriented Event Clustering (ROEC) procedure that calibrates the student model distribution with gold- standard event clusters while drawing distant supervision from the previously-generated event-specific rationales;
    \item We train the student model along with a frozen teacher distribution to model coreference probabilities of sampled event pairs. 
\end{enumerate*}

Fig.~\ref{fig:full_pipeline} provides a schematic of our approach. In this paper, we use LLaMA 2-7B-Chat~\cite{touvron2023llama} for the teacher model and Longformer-base~\cite{beltagy2020longformer} for the student model.

\subsection{Datasets}
We evaluate our method across three English event CDCR corpora with varying levels of referential ambiguity and difficulty.

\paragraph{Event Coreference Bank Plus (ECB+)}
Most prior works in event CDCR have evaluated on ECB+~\cite{cybulska2014using} due to the wide variety of topics that it covers. The lexica used in different ECB+ topics are largely distinct, leading to less overall ambiguity~\cite{bugert2020breaking}. We follow \citet{cybulska-vossen-2015-translating}'s approach for training, validation and testing splits. 

 \paragraph{Gun Violence Corpus (GVC)}
The GVC~\cite{vossen2018don} contains annotated events specifically in the domain of gun violence. The similar lexicon across event mentions leads to a high referential ambiguity. This tends toward coreference chains with a more realistic (i.e., non-Zipfian) distribution of events in text descriptions. This makes GVC more challenging especially in a CDCR setting. We use the splits from \citet{bugert2021crossdocument}. 

\paragraph{AIDA Phase 1}
The AIDA Phase 1 corpus \cite{tracey2022study} consists specifically of events in the domain of the Russia-Ukraine conflict, which are annotated based on their potential for conflicting perspectives.\footnote{The data is available from the Linguistic Data Consortium under catalog number LDC2019E77.}
This corpus's test set is larger than its training set, which makes it additionally challenging. To the best of our knowledge, the only evaluation to date performed on this dataset was performed by members of our team~\cite{nath2024multimodal}. As before, we follow the splits from \citet{tracey2022study}.
 

\subsection{Event Coreference Rationale Generation}

We define rationale generation with LLMs as an abductive reasoning problem~\cite{paul1993approaches}. Given a pair of event mentions and their contexts as an \textit{observation} ($e_1$ and $e_2$) and the coreference gold label as an \textit{outcome} ($g$), the LLM should generate the most probable \textit{hypothesis} or \textit{rationale} ($r^*)$, where:

\begin{equation}
 \small
r^*=\arg \max _{r^i} P\left(R=r^i \mid e_1, e_2, g\right)
\label{eq:rationale-1}
\end{equation}

We assume mutual exclusivity of rationales \cite{gordon2017formal}, such that one plausible rationale automatically excludes other rationales for an event mention pair.  This allows us to rewrite Equation \ref{eq:rationale-1} using Bayes Rule conditioned on the gold standard ($g$) as:
\begin{equation}
\small
   P\left(R \mid e_1, e_2, g\right) \propto P\left(g \mid e_1, e_2, R\right) \cdot P\left(R \mid e_1, e_2\right)
\label{eq:rationale-2}
\end{equation}

Since the gold coreference labels are meant to be the ground truth, we have:
 \vspace{-2mm}
\begin{equation}
\small
   P\left(R \mid e_1, e_2, g\right) \propto P\left(R \mid e_1, e_2\right)
\label{eq:rationale-3}
\end{equation}

We use gold coreference labels in our prompts to guide rationale generation. This establishes a one-to-one map\footnote{Coreference annotation typically relies on a structured knowledge base to minimize this set of plausible hypotheses~\cite{vossen2018don,tracey2022study}.} between an event mention pair and its rationale while also reducing the cost (computational or otherwise) associated with additional hypothesis generation.

\begin{figure}[t]
  \includegraphics[width=0.5\textwidth]{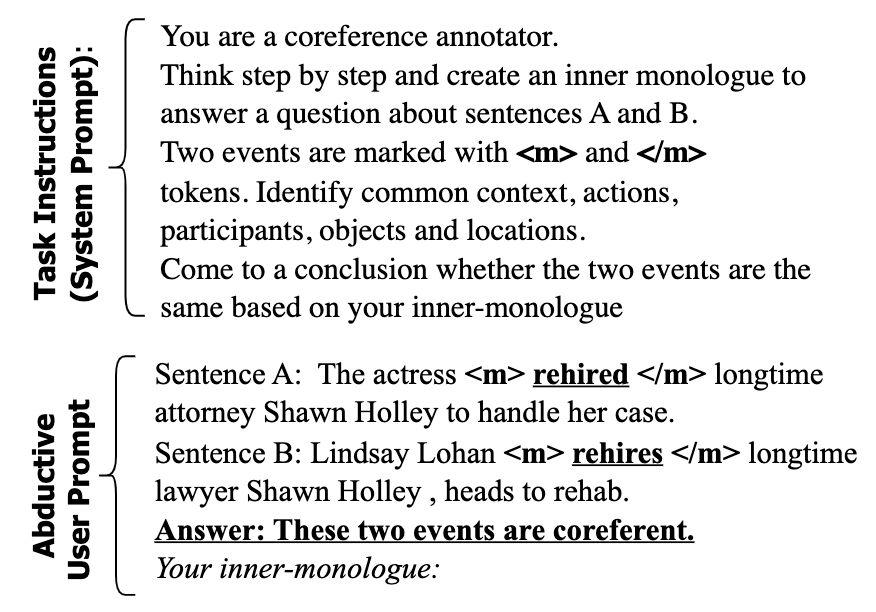}
\caption{Prompt format for inner monologue-based FTR generation conditioned on the gold label (underlined). \texttt{<m>} and \texttt{</m>} demarcate the event triggers.}  \label{fig:inner_monologue_prompt}
\end{figure}

Rationales are generated with LLaMA 2-7B-Chat~\cite{touvron2023llama}, a foundation model fine-tuned for "human-like" conversation. LLaMA 2's open-weight nature means the model distribution remains accessible. Since the model is otherwise not fine-tuned for event coreference, we prompt the language model to ground its rationale to event coreference-specific arguments such as participants, times, entities, and locations. This provides event-specific context that grounds the output to the event mention pair. This assures mutual exclusivity and provides information that can be used in representational learning techniques \cite{murty-etal-2020-expbert,kenyon2018resolving} for aligning or "interpreting" an event mention within the context of a rationale. For sampling, we use a temperature parameter of 0.7 for randomness and a top-$p$ of 0.9 to ensure diversity in the tokens. We constrain  generation to a maximum of 512 tokens. See Fig.~\ref{fig:inner_monologue_prompt} for prompt formatting, and Table~\ref{tab:rationale_stats} for our rationale dataset statistics.  These rationales or intermediate reasoning steps are then used to guide the other two stages of our procedure (Secs.~\ref{ssec:roec} \& \ref{ssec:kd}).

\subsection{Descriptive Statistics of Generated Rationales}
\label{app:rationale-stats}

\begin{table}[htb]
    \centering
    \resizebox{0.90\linewidth}{!}{
    \begin{tabular}{@{}llllcccc@{}}
        \toprule
        & \multirow{2}{*}{} & \multicolumn{3}{c}{Corpus} \\
        \cmidrule{3-7}
        && ECB+ & GVC & AIDA  \\
        \midrule
        & \# Event Pairs & 41,334 & 80,060 & 17,306 \\
        & \# Total Tokens & 12.3M & 24.5M  & 5.3M  \\
        & \# Unique Tokens & 12.2k & 13.0k & 7.5k \\
        & Avg. Token Length & 4.7 & 4.6 & 4.8 \\
        & Avg. Tokens/FTR & 300 & 305 & 310 \\
        & Self-BLEU & .77 (.66) & .82 (.75) & .79 (.78) \\

        \bottomrule
    \end{tabular}
    }
    \caption{Descriptive statistics of generated inner monologue-based free text rationales (FTRs) for ECB+, GVC, and AIDA Phase 1 at the corpus level. Self-BLEU scores of the gold coreference mentions are shown within parentheses.}
    \label{tab:rationale_stats}
\end{table}

In order to assess the diversity and the uniqueness of the generated rationales, we also conducted a lexical diversity analysis at the token and $n$-gram level. LLM-generated explanations typically tend to be less diverse and unique compared to human-written examples~\cite{welleck2019neural, west-etal-2022-symbolic}. We do not have a sample of human-generated rationales to compare to, but while a token-level diversity can provide an estimate of the exact lexical distribution in the rationales, a "softer" estimation can be drawn using the Self-BLEU~\cite{zhu2018texygen} metric. This gives us an idea of how similar rationales are amongst themselves at the $n$-gram level (higher value means more similar).

We report Self-BLEU scores for generated rationales for each corpus using a 10\% random sampling with a fixed seed. Table~\ref{tab:rationale_stats} shows that lexical uniqueness (unique tokens) is likely a function of the overall lexical distinctness of the corpus and not directly proportional to the cardinality of the full token set. For instance, ECB+ has almost twice the proportion of unique tokens to total token count when compared to GVC. Additionally, Self-BLEU scores of the corresponding gold coreference mentions (drawn directly from the corpora) are generally lower than those of FTR sets. More soft-uniqueness or $n$-gram diversity in the gold mentions could be due to the step-by-step nature of abductive reasoning where multiple angles of reasoning, and therefore wording, using can co-occur with similar context. We also conducted a human evaluation of the FTRs across various markers of fidelity and quality motivated by \cite{wiegreffe-etal-2022-reframing}. Appendix~\ref{app:im-quality} contains details of the human evaluation component. 


\subsection{Rationale Oriented Event Clustering (ROEC)}
\label{ssec:roec}


We perform "Rationale-Oriented Event Clustering" (ROEC) by directly optimizing the student model to encode cluster-level information in the coreference graph. Since the generated rationales come from a disparate model distribution, we align event pairs with their corresponding rationales in the student model's latent space.

 
Positive training samples consist of all mention pairs belonging to the same coreference cluster. When collecting negative samples, we want to avoid overwhelming the training distribution with non-coreferent pairs, which constitute most of any CDCR corpus. Since a rationale is assumed to be a supporting hypothesis for a specific event mention pair, as opposed to any other pair or at the cluster level, minimizing the search space is crucial, especially without a document-clustering step. As such, we sample negative pairs using the non-oracle heuristic from \citet{ahmed20232} with a low-threshold of 0.05. This heuristic, depending on the threshold, selectively retrieves hard negative pairs using a cache of synonymous lemma pairs occurring in the coreference chains. Since ECB+ and GVC contain almost no inter-cluster coreferents \cite{held-etal-2021-focus}, this makes the process of generating negative samples tractable and avoids non-informative inter-cluster candidates.

We encode an event mention pair ($e1$, $e2$) into a common representation $\overrightarrow{p}$ by extracting the {\tt [CLS]} token representation of the concatenated input from the last hidden layer of the student model. We encode the supporting rationale $\overrightarrow{r}$ similarly in the student. For training, we use a joint-optimization framework: a cross-entropy based cluster loss ($\mathcal{L}_{\text{cluster}}$) (\ref{eq:roec-1}) to predict the cluster label of the event pairs, and a cosine-distance based loss ($\mathcal{L}_{\text{rationale}}$) (\ref{eq:roec-2}) with a tuned weight penalty ($\lambda$) to align the vector embeddings for the rationales to those of the corresponding event pairs. For training batch size $m$, total number of clusters\footnote{Following \citet{kenyon2018resolving}, we cluster all singletons into a single dummy label. Following~\citet{rahman-ng-2009-supervised}, negative pairs are assigned to this $N+1^\text{th}$ class in $\mathcal{L}_{\text{cluster}}$ (\ref{eq:roec-1}).} $N$, predicted cluster probabilities $\hat{y}_i$, and ground truth cluster vector ${y}_i$, we minimize a joint-optimization loss given by Equation \ref{eq:roec-3}, with components (\ref{eq:roec-1}) and (\ref{eq:roec-2}):

 
 
 \begin{equation}
\small
 \mathcal{L}_{\text {cluster}}= -\frac{1}{m} \sum_{i=1}^{m} \sum_{n=1}^{N+1} y_i \cdot \log(\hat{y}_i)
 \label{eq:roec-1}
 \end{equation}

  \begin{equation}
\small
   \mathcal{L}_{\text {rationale}} \left(\overrightarrow{p}, \overrightarrow{r}\right)=\sum_{i=1}^{m} \left(1-\frac{\overrightarrow{p_i} \cdot \overrightarrow{r_i}}{\left\|\overrightarrow{p_i}\right\|\left\|\overrightarrow{r_i}\right\|}\right)
 \label{eq:roec-2}
 \end{equation}
 
  \begin{equation}
 \small
   \mathbf{L}=\mathbf{L}_{\text{cluster}}+\lambda \mathbf{L}_{\text{rationale}}
 \label{eq:roec-3}
 \end{equation}

 
\subsection{Coreference Knowledge Distillation}
\label{ssec:kd}


We use the optimized student model from the previous step as the backbone encoder. Our classifier then learns coreference probabilities of event pairs by optimizing the task-supervision component against gold labels while simultaneously aligning the student distribution with the teacher model representations of the rationales as soft labels. 
We optimize a "knowledge distillation loss" given by Equation \ref{eq:kd-loss}.

  \begin{equation}
 \small
\mathcal{L}_{\text {KD}} = \mathcal{L}_{\text {task}} + \lambda_{1} \mathcal{L}_{\text {attention}} + \lambda_{2}\mathcal{L}_{\text {pooled}}
 \label{eq:kd-loss}
 \end{equation}
 
We estimate the regularization parameters with a grid search over the validation set. We find $\lambda_{1}$ = 1 and $\lambda_{2}$ = 0.01 to work best for student model convergence. Each individual component is defined as follows.

\paragraph{Task Component}


This uses a pairwise scorer framework~\cite{caciularu2021cdlm} to train the student model. Document pairs containing the individual event-trigger spans are encoded in the student model into a common representation that consists of the {\tt [CLS]} representation of the document pair and the attention components of representations $e_1$ and $e_2$ as well as of Hadamard product $e_1 \odot e_2$. This common representation is then fed into the classification layer (multi-layer perceptron) of the student model to estimate coreference probabilities. For this task-specific supervision, we minimize the Binary Cross Entropy Loss, $\mathcal{L}_{\text{task}}$:
 \vspace{-2mm}
\begin{equation}
 \small
    \mathcal{L}_{\text{task}} =-\frac{1}{m} \sum_{i=1}^{m}\left(y_i \cdot \log \hat{y}_i+\left(1-y_i\right) \cdot \log \left(1-\hat{y}_i\right)\right)
\end{equation}
where $y$ and $\hat{y}$ are the true and predicted coreference probabilities in a sample batch of size $m$.

\paragraph{Distillation with Rationales}

For the distant supervision-based distillation, aligning the student ($S$) and teacher ($T$) distributions is carried out with the rationales ($\boldsymbol{R}$) encoded across the attention states ($\boldsymbol{R}_{a}^*$) and the last hidden states ($\boldsymbol{R}_{h}^*$).


\subparagraph{Attention Loss}
Transformer-based language models like BERT tend to capture high-level linguistic knowledge, including coreference signals, in their attention states, distributed across the various heads~\cite{clark2019does}. To align the attention states, we minimize the squared $L^2$ norm between the last-layer attention representations of the rationales as encoded in the student ($\boldsymbol{R}_{a}^S$) and teacher ($\boldsymbol{R}_{a}^T$). Motivated by \citet{jiao2020tinybert}, we apply a mapping function ($f(i) = i + H - h;\quad 0 < i \leq h$) from student attention head indices to teacher heads, where the $i^{\text{th}}$ student head sources supervision from the $f(i)^{\text{th}}$ teacher head. $H$ and $h$ represent the total number of attention heads in the teacher and the student, respectively. For instance, the first student attention head is mapped to teacher attention head ($1$ + $H$ - $h$).  Therefore,


\begin{equation}
\small
    \mathcal{L}_{\text {attention}} = \sum_{i=1}^h\|\boldsymbol{R}_{a_i}^S-\boldsymbol{R}_{a_{f(i)}}^T\|_2^2
\end{equation}


\subparagraph{Hidden-state Loss}
Similarly, here we extract the final-layer pooled rationale representations from the student and teacher, and minimize the squared $L^2$ norm between them. A learnable linear projection matrix $\boldsymbol{W}_{T\xrightarrow{}{S}}$ is used to project the teacher's 4096D hidden representation into the student's 768D latent space, resulting in:

\begin{equation}
\small
  \mathcal{L}_{\text {pooled}} = \|\boldsymbol{R}_{h}^S - \boldsymbol{R}_{h}^T\boldsymbol{W}_{T\xrightarrow{}{S}}\|_2^2
  \end{equation}


\section{Experiments}
\paragraph{Ablations}

We evaluate four different variants of our model, to establish how each component contributes to final performance. 
\begin{enumerate*}[label=\arabic*)]
    \item $Long_\text{\normalfont paired}$ establishes a baseline using no ROEC or knowledge distillation. This resembles traditional representational learning systems that leverage natural language rationales. We implement \citet{murty-etal-2020-expbert}'s method and "pair" rationales with the corresponding event mentions by extracting the {\tt [CLS]} token representation from the pairwise scorer framework which is trained using a simple BCE loss.
    \item $Long_\text{\normalfont+ROEC,-KD}$ includes ROEC optimization with additional training with a task-specific BCE loss ($\mathcal{L}_{\text{task}}$).
    \item $Long_\text{\normalfont -ROEC,+KD}$ excludes ROEC but includes coreference knowledge distillation.
    \item $Long_\text{\normalfont+ROEC,+KD}$ uses both components.
\end{enumerate*}

\paragraph{Training Parameters}

For training the ROEC phase (Sec.~\ref{ssec:roec}), we use an Adam optimizer~\cite{kingma2014adam} with a batch size of 40 and a model learning rate of $1e-5$ for 20 epochs. For training coreference knowledge distillation (Sec.~\ref{ssec:kd}) we use a smaller batch size of 16 to ensure optimal performance. We use a model learning rate of $1e-5$ and a classifier learning rate of $1e-3$ and train for 10 epochs. We use a single NVIDIA A100 GPU for training both phases. ROEC and coreference knowledge distillation take roughly 20 minutes and 45 minutes, respectively, for a training a single epoch.


\paragraph{Inference and Evaluation}

We follow a simple connected components-based clustering approach at inference to generate coreference chains. {\it No FTRs are included in the input to the model at inference time.} We score candidate pairs with only the gold coreference labels using our coreference knowledge-enhanced model (Sec.~\ref{ssec:kd}). These scores are then used to construct an affinity graph to identify connected components using a threshold of 0.5. Thereafter, the generated coreference chains are evaluated against the gold clusters to calculate final coreference metrics. Following \citet{held-etal-2021-focus}, we focus on the $B^3$ metric, which is sensitive to incorrectly clustered singletons.

\section{Results}


Table~\ref{tab:bcub_alldata} shows evaluation results over the ECB+, GVC, and AIDA test sets. Results provided are single runs after robust hyperparameter tuning on the validation sets. We compare our method (with ablations) to relevant previous baselines, focusing on the $B^3$ metric as mentioned above. Since we preprocess according to \citet{ahmed20232}'s heuristic that samples from a cache of coreferent lemma pairs built only over the training set (see Sec.~\ref{sec:related}), we compare to their result reported using this heuristic. Appendix~\ref{app:results} shows results according to other common coreference metrics.

To demonstrate that a generative LLM alone does not perform equivalently, we compare to zero-shot results from LLaMA 2-7B-Chat (our teacher model) and GPT-3.5-Turbo. These models were prompted to provide a single-word ({\it yes}/{\it no}) answer to whether or not the events given in the prompt are coreferent. Prompt format is a slight variant of that given for rationale generation in Fig.~\ref{fig:inner_monologue_prompt}. See Appendix~\ref{app:prompt} for more details.



\begin{table}[htb]
    \centering
    \resizebox{0.90\linewidth}{!}{
    \begin{tabular}{@{}llllcccc@{}}
    \toprule
        &&& \multicolumn{3}{r}{$B^3$} \\
        \cmidrule{4-8}
        & Methods &&  ECB+ && GVC && AIDA  \\
        \midrule
          & \citet{bugert2021crossdocument} && - && 59.4 && - \\
         & \citet{cattan2021cross} && 81.0 && - && - \\
         & \citet{caciularu2021cdlm} && 85.6 && - && - \\
         & \citet{held-etal-2021-focus} &&  85.7 && 83.7 && - \\
         & \citet{ahmed20232} && 82.4 && 77.7 && - \\
         
\cdashline{1-8}
        &LLaMA 2-7B-Chat && 77.7 && 53.5 && 47.8\\
        & GPT-3.5-Turbo && 79.8  && 49.6 && 56.0 \\
        & $Long_\text{paired}$ && 81.8  && 75.3 && 58.1 \\
        & $Long_\text{+ROEC,-KD}$ && 85.9 && 80.6 && 61.2 \\ 
        & $Long_\text{-ROEC,+KD}$ && 84.4 && 82.5 &&  61.5 \\
        &  $Long_\text{+ROEC,+KD}$ && \textbf{86.8} &&   \textbf{84.3} &&  \textbf{64.5} \\
        
    \bottomrule
    \end{tabular}
    }
    \caption{$B^3$ $F_1$ results using our coreference knowledge distillation framework on the ECB+, GVC and the AIDA Phase 1 test sets.}
    \label{tab:bcub_alldata}
    \vspace*{-3.5mm}
\end{table}

Our best model, using both ROEC and knowledge distillation, outperforms previous baselines on both ECB+ (+1.1 $B^3$ $F_1$) and GVC (+0.6 $B^3$ $F_1$), as well as zero-shot LLaMA 2-7B-Chat and GPT-3.5-Turbo. AIDA Phase 1 is a new, challenging dataset, where we establish a new baseline.

$Long_\text{+ROEC,+KD}$ learns from the teacher distribution, but we find it significantly outperforms both the zero-shot teacher model (LLaMA 2-7B-Chat) and the much larger GPT-3.5-Turbo.
Without task-specific finetuning, general purpose models tend to under-perform in reasoning-based tasks compared to smaller fine-tuned variants~\cite{ho2022large}, but our results suggest that the teacher-generated rationales contain useful information for coreference decisions, such that when they are conditioned on the gold label, the distilled knowledge optimizes the task-based encoder for better performance, despite such larger models performing poorly in a zero-shot setting~\cite{wiegreffe-etal-2022-reframing}.

\subsection{Ablation Tests}

When we ablate the knowledge distillation component, we find that adding KD to a model that only performs ROEC ($Long_\text{+ROEC,-KD}$) boosts performance by +0.9 $B^3$ $F_1$ (ECB+), +3.7 $B^3$ $F_1$ (GVC) and +3.3 $B^3$ $F_1$ (AIDA). This further attests to the informativeness of the generated FTRs.

Performing ROEC alone results in a 3--5 $B^3$ $F_1$ performance boost compared to the simple paired representation learning approach ($Long_\text{paired}$). Adding KD on top of that boosts performance more on the GVC and AIDA corpora than on ECB+. This suggests that the knowledge distillation component that sources task-specific supervision from the teacher brings an additional performance gain when there is more referential ambiguity, as in GVC, or conflicting event descriptions, as in AIDA Phase 1. In ECB+, topics are lexically distinct which results in rationales for a given topic having a similar lexical distribution. This suggests that when ROEC alone boosts performance more than KD alone, it is likely due to the nature of the corpus. This is consistent with \citet{bugert2021crossdocument}'s observation that CDCR systems with document clustering tend to overfit to the structure and link distribution of ECB+ largely due to its lexical distinctions between topics.

\section{Discussion}
\label{sec:disc}

We ran \citet{ahmed20232}'s pipeline over the same data and compared where their method and ours make different decisions on ECB+ and GVC samples. Since we use the same preprocessing here, this compares the performance of the two pairwise discriminators used: plain Longformer vs. our knowledge-enhanced version.

\begin{table}[htb]
    \centering
    \resizebox{ \linewidth}{!}{
    \begin{tabular}{@{}llcccc@{}}
        \toprule
        & \multirow{2}{*}{} & \multicolumn{2}{c} {ECB+} & \multicolumn{2}{c} {GVC} \\
        \cmidrule(lr){3-4} \cmidrule(lr){5-6} 
        && $Long_\text{+ROEC,+KD}$ & $Long$ & $Long_\text{+ROEC,+KD}$ & $Long$ \\
        \midrule

        & \# pos. & 825 (3506) & 11 (3506) & 470 (2633) & 20 (2633) \\
        & \# neg. & 483 (2041) & 334 (2041) & 428 (6768) & 70 (6768)  \\
     
        \bottomrule
    \end{tabular}
    }
    \caption{Number of positive (coreferent) and negative samples, per dataset, on which the indicated model succeeded and the other failed. Total number of pairs of the given label in each dataset is shown in parentheses.}
    \label{tab:lh_comparison}
\end{table}

Table~\ref{tab:lh_comparison} shows the comparative error analysis. $Long_\text{+ROEC,+KD}$ succeeds at linking coreferent ECB+ pairs that plain Longformer fails on 75x as often as the reverse (825 to 11). The margin is 23.5x on GVC. While the margin is lower on non-coreferring samples, $Long_\text{+ROEC,+KD}$ is still a substantially stronger performer than plain Longformer here as well.

\begin{figure}[t]
  \includegraphics[width=0.5\textwidth]{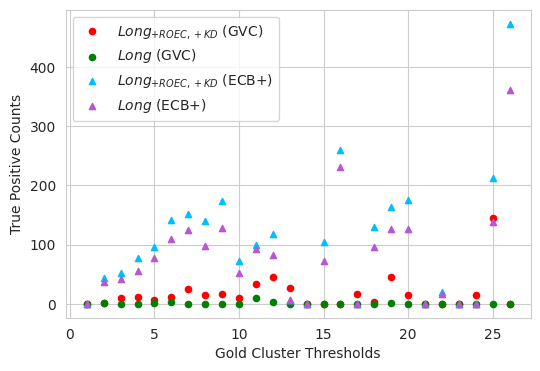}
\caption{Distribution of mentions correctly resolved by the indicated model vs. cluster size.}  \label{fig:cluster_plot}
 \vspace*{-2mm}
\end{figure}

Fig.~\ref{fig:cluster_plot} shows the distribution of mentions correctly resolved by the indicated model as a function of cluster size in the two corpora. $Long_\text{+ROEC,+KD}$ is nearly globally more successful at making coreference links, and disparities grow as cluster size increases.  This effect is seen in both corpora but is particularly pronounced in ECB+.




Using knowledge distilled from FTRs, we are able to make more correct coreference decisions than plain Longformer, even without using the rationale in the input at inference time. This holds true across both ECB+ and GVC test sets. Our method also establishes a new baseline for the AIDA Phase 1 corpus. 

\paragraph{Qualitative Analysis}

\begin{table*}[h]
\resizebox{\textwidth}{!}{
    \footnotesize  
    
    \begin{tabular}{llp{8cm}p{8cm}}
    \toprule
    & \textbf{True Label} & \textbf{Document A} & \textbf{Document B} \\
    \midrule
    (a) & {\it Not coreferent}  & 6.1-magnitude quake \underline{strikes} Indonesia's Aceh province & 6.1-magnitude quake strikes Indonesia's Aceh province in region \underline{hit} by 2004 tsunami \\
    & \multicolumn{3}{p{18.5cm}}{\textbf{FTR sample}: \textit{"In Document A, the earthquake is the primary focus, while in Document B, the earthquake is mentioned in the context of a tsunami that occurred in 2004."}} \\
    \midrule
    (b) & {\it Coreferent}  & Advanced Micro Devices (NYSE: AMD) \underline{announced} the largest acquisition in its history Monday, paying about \$5.4 billion to acquire graphics chip specialist ATI Technologies. & RBC Capital Markets downgraded ATI Technologies to "sector perform" from "outperform" after Advanced Micro Devices \underline{announced} its intention to buy the Canadian graphics chip-maker for \$5.4 billion on Monday. \\
    & \multicolumn{3}{p{18.5cm}}{\textbf{FTR sample}: \textit{"[...] I see the phrase `paying about \$5.4 billion to acquire graphics chip specialist ATI Technologies...' Oh, so it's ATI Technologies that AMD acquired! Fast forward to Document B, and I see the same phrase `Advanced Micro Devices <m> announced </m> its intention to buy the Canadian graphics chip-maker for \$5.4 billion on Monday...' Wait, what?! Didn't we just establish that AMD acquired ATI Technologies?"}} \\
    \midrule
    (c) & {\it Not coreferent}  & D'kai Vanlandingham, 17, of Northeast Baltimore, was \underline{shot} and killed just after 10 p.m. Saturday on the 3400 block of Juneway in Belair-Edison, police said. & Martin Ibarra, 35, of South Baltimore, was \underline{shot} and killed shortly before 10 p.m. Friday on the 1400 block of Ramsay St. in New Southwest-Mount Clare, police said. \\
    & \multicolumn{3}{p{18.5cm}}{\textbf{FTR sample}: \textit{"In document A, the victim is identified as D'kai Vanlandingham, while in document B, the victim is named Martin Ibarra. [...] Document A states that the incident happened on the 3400 block of Juneway in Belair-Edison, while document B mentions the 1400 block of Ramsay St. in New Southwest-Mount Clare. [...] [T]here are two separate incidents involving people being shot and killed in different locations in Baltimore."}} \\
    \midrule 
    (d) & {\it Coreferent}  & Two suspects were arrested Monday in the killing of a 38-year-old man who died more than two weeks after he and four others were \underline{wounded} in a shooting outside a Little Rock home. & Five people, including Duhart, were \underline{injured}. The other victims involved in the shooting were identified as: \\
    & \multicolumn{3}{p{18.5cm}}{\textbf{FTR sample}: \textit{"[...] Document A is clear and specific [...]: The event occurred outside a Little Rock home. One person was killed, and four others were injured. The event involved a shooting. From Document B, I know that five people, including Duhart, were injured [...]."}} \\
    \midrule
    \bottomrule
    \end{tabular}}
    \caption{Test samples from ECB+ (a--b) and GVC (c--d) that our method ($Long_\text{+ROEC,+KD}$) links correctly and plain Longformer (reproduced according to \citet{ahmed20232}) fails on. Mention triggers are underlined. As these are test samples, the presented abridged FTR samples are not included in the input for inference, but are given as examples of the kinds of information our model is likely to incorporate in making its decision.}
     \label{tab:error-analysis}
\end{table*}

Table~\ref{tab:error-analysis} presents two test samples from each dataset that our best model clustered successfully and plain Longformer did not. We present snippets of the FTRs generated for these samples as well. As FTRs were not included in the input to the model during inference, these simply serve to provide an additional way of interpreting the results, by showcasing the kind of information contained within an FTR that, when distilled into the student model, plausibly contributes to a correct coreference linkage:

\begin{enumerate}[label=(\alph*)]
    \item Although both documents refer to a 6.1 magnitude earthquake in Aceh, the mention trigger in Document B actually refers to the 2004 tsunami that hit the same region.  The FTR mentions this distinction in context.
    \item The two mentions contain the same elements (organizations, prices) but in a different order. The FTR correctly ascribes the organizations and the \$5.4 million price to both documents.
    \item The two events take place in the same city and the mentions actually have identical syntax. However, the FTR correctly identifies that they refer to different people and regions/neighborhoods of Baltimore.
    \item Only Document A mentions the location while only Document B mentions the name of the wounded/deceased, but the FTR is able to note that in one document there were 5 victims while in the other there were 1+4 victims.
\end{enumerate}

This shows that in areas like exposing alignment or divergence between named entities, temporal contexts, or syntax across mention pairs, FTRs are providing useful information by making explicit what may be implicit in the raw text. Our knowledge distillation procedure is able to take this information from the attention heads and hidden states of the teacher model and project it into the student model for finer-grained coreference decisions.

\section{Conclusion and Future Work}
\label{sec:conc}

In this paper, we presented a novel event CDCR technique that used free-text rationales from a generative LLM to provide additional supervisory signals for coreference. We accomplished this through a combination of clustering rationale representations with the corresponding event mentions in a student model's space, and by distilling information from the generative teacher model into the smaller student. We achieved SOTA $B^3$ scores on the ECB+ and GVC benchmarks, and established novel benchmark performance on the challenging AIDA Phase 1 dataset. We also ablated the contributions of different components of our model and examined the kinds of information our model is likely leveraging toward its performance.

It would be too strong a claim to say that the generated FTRs consistently display coherent reasoning in and of themselves (for example, sample (b) in Table~\ref{tab:error-analysis} seems to treat the two mentions as events occurring in sequence rather than to be considered in parallel; this may be an artifact of the autoregressive generation mechanism), but by forcing the LLM to explicitly generate output concerning the common entities and arguments of the events in question, it is also forced to generate information more explicitly relevant to coreference links than just the event triggers themselves and immediately surrounding context. This resembles a "dense paraphrasing"-like procedure~\cite{tu-etal-2023-dense} that provides enriched event descriptions that we then distill into our model using ROEC and knowledge distillation from the teacher distribution.

Our results demonstrate that although imperfect, automatically-generated rationales for event coreference contain useful information toward the decision. Using AI-generated rationales as soft-labels might prove useful as a way to decrease annotator workload in cognitively-heavy tasks like annotating coreference resolution corpora~\cite{zhao-etal-2023-cross}.

This opens the way to future work to improve the utility of FTRs. For example, filtering methods may be used to exclude FTRs with lower-quality coreference knowledge. Techniques such as~\citet{west-etal-2022-symbolic} that employ a separate but smaller critic model can be trained on a small sub-sample of high-quality rationales written by trained coreference annotators, to further enhance coreference-specific knowledge distillation with lower annotation expenses. FTRs generated using a more powerful model like GPT-4 could also be beneficial in extending extant CDCR corpora with more explicit soft-labels and can likely enhance systems that need to detect cross-subtopic coreference in corpora such as FCC~\cite{bugert2020breaking}, albeit at the cost of accessibility to the source model. FTRs with validated gold cluster-level information could be leveraged especially in preclustering to reduce cross-computations, making such systems more generalizable. 







\section*{Limitations}

While our results demonstrate that automatically-generated rationales for event coreference contain useful information toward the decision, there remains the fact that like all current generative AI models, LLaMA 2-7B-Chat may "hallucinate" or output fallacious information.  For instance, for a GVC mention pair A: "\textit{3-year-old shot, \underline{killed} in Stockon while riding in car.}" and B: "\textit{The girl, identified as Melanie Martinez of Stockton, was the only person in the vehicle who was hit by the shots, and her family drove her straight to a nearby hospital, according to police. She was pronounced  \underline{dead} at the hospital.}", the generated FTR mentions that "\textit{Both documents mention the girl was pronounced dead at a hospital.}"  However, this fact was only mentioned in the second document. We designed our study with the goal of minimizing such occurrences (see Appendix~\ref{app:design_choices}), but the frequency and effect of such hallucinations remains to be investigated.

On the zero-shot results comparisons, there may be some sensitivity to the prompt given to the two competing models. Some surface-level prompt engineering was conducted to ensure that the models only provided one word answers in the zero-shot setting and could make a coreference distinction in clear cases, in order to provide a reasonable and evaluable baseline comparison. As the focus of this paper is not on prompt engineering for closed models, we did not investigate this further.

Our results should be considered in the context of the entire pipeline.  There are many ways to preprocess CDCR corpora to render the task more tractable.  We eschewed the document clustering step of many popular methods due to computational expense and tendency to exclude many valid inter-cluster links (see Sec.~\ref{sec:related}). That left us two filtering strategies from recent work: \citet{held-etal-2021-focus}'s discourse modeling in the latent space and \citet{ahmed20232}'s heuristic filtering. \citet{ahmed20232}'s method is faster, so we used this preprocessing step. Despite this, we were still able to exceed \citet{held-etal-2021-focus}'s $B^3$ $F_1$ scores on ECB+ and GVC.

Given the nature of their subject material, the Gun Violence and AIDA Phase 1 corpora may be troubling to some, including, apparently, a generative LLM.  For 167 of 7,314 ($\sim$2.28\%) of AIDA Phase 1 samples and 80 out of 10,355 ($\sim$0.77\%) of GVC, LLaMA 2-7B-Chat would not generate a definite answer for the event pair when evaluated in the zero shot setting, citing ethical and moral standards and the fact that the event mentions (and therefore prompt) contained descriptions of violence and harm. These samples had to be discarded from evaluation. This effect was not observed when using LLaMA 2-7B-Chat to generate the FTRs for training our methods.

\section*{Acknowledgments}

Okay, let's do this! We'd like to thank the evaluators of our free-text rationale examples: Nada Alalyani, Jack Fitzgerald, Rahul Ghosh, Anju Gopinath, Huma Jamil, Changsoo Jung, Ibrahim Khebour, and Hannah VanderHoeven. Our thanks also go out to the anonymous reviewers whose comments helped improve the final version of this paper.

\bibliography{anthology,custom,abductive_reasoning}

\appendix

\section{Package and Pre-/postprocessing Details}
\label{app:packages}

We use the pretrained Longformer-base weights as they appear in the HuggingFace library.\footnote{\url{https://huggingface.co/allenai/longformer-base-4096}} For LLaMA 2-7B-Chat, we use the downloaded weights from Meta\footnote{\url{https://ai.meta.com/llama/}} which were then converted to the HuggingFace Transformers format for their pretrained libraries. For zero-shot evaluation using GPT-3.5-Turbo, we use OpenAI's {\tt completions} API gateway.\footnote{\url{https://platform.openai.com/docs/guides/text-generation/completions-api}} We use the NLTK library\footnote{\url{https://www.nltk.org/api/nltk.tokenize.html}} for tokenization when generating the rationale statistics at the token-level. For lemma-based heuristic candidate event generation, we use the popular spaCy Lemmatizer pipelines\footnote{\url{https://spacy.io/api/lemmatizer}}. For getting the final coreference clusters after creating the affinity graph (post-transitive closure), we use the CoVal coreference scorer~\cite{moosavi2019minimum}.

\section{Further Details on Motivation for Design Choices}
\label{app:design_choices}

\paragraph{Exclusion of repulsive regularization during ROEC}
Since our rationales are abductive in nature and have been conditioned on the gold coreference labels, it is likely that the step-by-step reasoning contains valid reasoning steps in support of the label, regardless of the actual status of the label (coreferent or not). Such informativeness of intermediate reasoning steps have also been observed in previous work~\cite{wiegreffe-etal-2022-reframing} that suggests that a general purpose LLM can still generate plausible explanations given a task, even if it displays sub-par performance on the task itself especially in zero-shot evaluations. 

Since our rationales incorporate multiple angles of reasoning about coreference, we hypothesize that the regularization process should only reward event representations that form similar clusters while events from separate clusters should remain unrewarded during training. This is because rationales for these events could still contain plausible hypotheses consistent with the cluster label. Therefore, we do not include a repulsive regularization component commonly used for creating separable clusters in pretraining event coreference models as seen in~\citet{kenyon2018resolving} and \citet{held-etal-2021-focus}.


\paragraph{Choice of temperate parameter}
The temperature we use for FTR generation (0.7) is the default value in LLaMA 2-7B-Chat. This creates controlled and focused responses to prompts without being fully deterministic. A conservative sampling strategy helps keep diversity of tokens in the rationales without added randomness that could negate the gold label or generate out-of-context outputs. Initial experimentation demonstrated 0.7 to be a reasonable value and a test of temperature values was determined to be out of scope for the paper.

\paragraph{Heuristic threshold optimization}
The threshold for the heuristic used to preprocess the data according to~\citet{ahmed20232}'s method is optimized using the validation data to minimize the loss of truly coreferent pairs, while pruning the large number of non-informative true negative mentions during candidate selection. A lower threshold also lets us select lexically misleading mention pairs that are actually not coreferent (hard negatives), that the heuristic fails on. These pairs are frequently seen in CDCR corpora, particularly in GVC~\cite{vossen2018don}. 

Since our rationales are defined as a one-to-one map with the corresponding mention pairs, such hard negatives paired with their rationales provide a richer signal during ROEC optimization. This helps the student model distinguish between pairs in the coreference cluster graph. Due to the sparsity of CDCR links~\cite{bugert2021generalizing}, the heuristic maintains a class balance between positive and negative pairs without resorting to a document clustering step. This allows pairwise classifiers to learn more efficiently from a relatively balanced class distribution.





\section{Additional Results Tables}
\label{app:results}

\begin{table*}[!ht]
    \centering
     
 \footnotesize
    \resizebox{\textwidth}{!}{
    \begin{tabular}{@{}lllrrrrrrrrrrrrrrrrr@{}}
    \toprule
    &&& \multicolumn{3}{c}{MUC} && \multicolumn{3}{@{}c@{}}{$B^3$} & & \multicolumn{3}{c}{$CEAFe$} && CoNLL\\
    \cmidrule{4-6} \cmidrule{8-10} \cmidrule{12-14} \cmidrule{16-16}
    &&& R & P & $F_1$ && R & P & $F_1$ && R &P & $F_1$ && \multicolumn{1}{r}{$F_1$}  \\ 
   \midrule

        
                


         & \citet{caciularu2021cdlm} && \textbf{87.1}  & 89.2 & \textbf{88.1} && 84.9 & 87.9 & 86.4 && 83.3 & 81.2 & 82.2 && 85.6 \\
        & \citet{held-etal-2021-focus} && 87.0  & 88.1 & 87.5 && 85.6 & 87.7 & 86.6 && 80.3 & \textbf{85.8} & 82.9 &&  85.7 \\
        &   \citet{yu2022pairwise} &&  88.1 & 85.1 & 86.6 && \textbf{86.1} & 84.7 & 85.4 && 79.6 & 83.1 & 81.3 && 84.4   \\
        & \citet{ahmed20232} (w/o oracle) && 80.0  & 87.3 & 83.5 && 79.6 & 85.4 & 82.4 && 83.1 & 75.5 & 79.1 && 81.7\\
\cdashline{1-20}

        &LLaMA 2-7B-Chat (zero-shot)  && 84.2  & 76.3 & 80.1 && 82.7 & 73.2 & 77.7 && 67.5 & 77.2 & 72.0 && 76.6  \\
        
       & GPT-3.5-Turbo (zero-shot)  && 81.7  & 81.0 & 81.4 && 81.0 & 78.6 & 79.8 && 76.1 & 77.0 & 76.5 && 79.2 \\

       & $Long_\text{paired}$ && 81.5  & 84.1 & 82.8 && 81.1 & 82.4 & 81.8 && 79.4 & 76.5 & 77.9  && 80.8 \\
& $Long_\text{+ROEC,-KD}$ (ours) && 79.4  & \textbf{92.4} & 85.4 && 79.8 & \textbf{93.1} & 85.9 && \textbf{89.1} & 76.1 & 82.1 && 84.5 \\ 
        & $Long_\text{-ROEC,+KD}$ (ours)  && 78.2  & 90.6 & 83.9 && 79.4 & 90.2 & 84.4 && 87.9 & 75.4 & 81.2 && 83.2 \\
        &  $Long_\text{+ROEC,+KD}$ (ours) && 84.1  & 92.0 & 87.9 && 82.4 & 91.7 & \textbf{86.8} && 88.9 & 80.5 & \textbf{84.5} && \textbf{86.4} \\


    \bottomrule
    \end{tabular}}
    \caption{ECB+ test set evaluation results}
    \label{tab:llama_ecb_full}
\end{table*}

\begin{table*}[!ht]
    \centering
   \footnotesize
    \resizebox{\textwidth}{!}{
    \begin{tabular}{@{}lllrrrrrrrrrrrrrrrrr@{}}
    \toprule
    &&& \multicolumn{3}{c}{MUC} && \multicolumn{3}{@{}c@{}}{$B^3$} & & \multicolumn{3}{c}{$CEAFe$} && CoNLL\\
    \cmidrule{4-6} \cmidrule{8-10} \cmidrule{12-14} \cmidrule{16-16}
    &&& R & P & $F_1$ && R & P & $F_1$ && R &P & $F_1$ && \multicolumn{1}{r}{$F_1$}  \\ 
   \midrule
         & \citet{bugert2021crossdocument} && 78.1 & 66.3 & 71.7&& 73.6 &49.9& 59.5 &&38.2& 60.9& 47.0 && 59.4\\
        & \citet{held-etal-2021-focus} && 91.8 &91.2 & 91.5 && 82.2 & \textbf{83.8} & 83.0 && 75.5 & \textbf{77.9} & \textbf{76.7} && \textbf{83.7} \\
        & \citet{ahmed20232} (w/o oracle) && 84.0  & 91.1 & 87.4 && 79.0 & 76.4 & 77.7 && 69.6 & 52.5 & 59.9 && 75.0 \\
        \cdashline{1-20}
        &LLaMA 2-7B-Chat (zero-shot) && \textbf{93.9}  & 84.3 & 88.8 && 89.5 & 38.1 & 53.4 && 28.9 & 54.9 & 37.9 && 60.0 \\
        & GPT-3.5-Turbo (zero-shot) && 88.6  & 81.9 & 85.1 && 82.6 & 35.4 & 49.6 && 27.1 & 41.1 & 32.7 && 55.8 \\
       & $Long_\text{paired}$ && 89.6  & 92.2 & 90.8 && 86.4 & 66.7 & 75.3 && 66.2 & 59.2 & 62.5 && 76.2 \\
        
        & $Long_\text{+ROEC,-KD}$ (ours) && 91.9  & 92.5 & 92.2 && \textbf{86.8} & 75.3 & 80.6 && 66.9 & 65.3 & 66.1 && 79.6 \\

        & $Long_\text{-ROEC,+KD}$ (ours) && 91.3  & 95.1 & \textbf{93.2} && 86.0 & 79.2 & 82.5 && \textbf{76.6} & 65.5 & 70.6 && 82.1 \\
        &  $Long_\text{+ROEC,+KD}$ (ours) && 91.6  & \textbf{94.2} & 92.9 && 86.7 & 82.1 & \textbf{84.3} && 75.8 & 68.1 & 71.7 && 83.0\\



    \bottomrule
    \end{tabular}}
    \caption{GVC test set evaluation results.}
    \label{tab:table_full_gvc}
\end{table*}

\begin{table*}[!ht]
    \centering
   \footnotesize
    \resizebox{\textwidth}{!}{
    \begin{tabular}{@{}lllrrrrrrrrrrrrrrrrr@{}}
    \toprule
    &&& \multicolumn{3}{c}{MUC} && \multicolumn{3}{@{}c@{}}{$B^3$} & & \multicolumn{3}{c}{$CEAFe$} && CoNLL\\
    \cmidrule{4-6} \cmidrule{8-10} \cmidrule{12-14} \cmidrule{16-16}
    &&& R & P & $F_1$ && R & P & $F_1$ && R &P & $F_1$ && \multicolumn{1}{r}{$F_1$}  \\ 
   \midrule 
        & LLaMA 2-7B-Chat (zero-shot) && \bf{69.1}  & 65.9 & 67.5 && \bf{60.0} & 39.7 & 47.8 && 45.7 & \bf{51.0} & 48.2 && 54.5 \\
        & GPT-3.5 Turbo (zero-shot) && 55.6  & 71.4 & 62.5 && 54.4 & 57.7 & 56.0 && 63.0 & 42.7 & 50.9 && 56.5 \\
       & $Long_\text{paired}$ && 57.5  & 76.4 & 65.6 && 54.4 & 62.4 & 58.1 && 68.0 & 44.4 & 53.7 && 59.1 \\
        
        & $Long_\text{+ROEC,-KD}$ (ours) && 56.9  & 87.7 & 69.1 && 52.3 & 73.7 & 61.2 && 83.1 & 47.6 & 60.5 && 63.6 \\
        & $Long_\text{-ROEC,+KD}$ (ours)  && 60.2  & 88.1 & 71.5 && 53.0 & 73.3 & 61.5 && 81.7 & 48.8 & 61.1 && 64.7 \\
        & $Long_\text{+ROEC,+KD}$ (ours)  && 60.6  & \bf{90.6} & \bf{72.6} && 53.3 & \bf{81.5} & \bf{64.5} && \bf{85.3} & 50.0 & \bf{63.0} && \textbf{66.7} \\



    \bottomrule
    \end{tabular}}
    \caption{AIDA Phase 1 test set evaluation results.}
    \label{tab:table_full_aida}
\end{table*}

In Tables~\ref{tab:llama_ecb_full}--\ref{tab:table_full_aida}, we present commonly used CDCR metrics~\cite{moosavi2019minimum}, comparing our systems' performances on ECB+, GVC, and AIDA Phase 1 to zero-shot evaluations, and previous baselines where available, to aid in future comparisons.  We show MUC~\cite{vilain1995model}, $B^{3}$~\cite{bagga1998algorithms}, $CEAF_e$, and CoNLL $F_1$ (the average of MUC, $B^{3}$ and $CEAF_e$ $F_1$) scores. Although we do not always beat previous evaluations on EBC+ and GVC on all metrics, we frequently either do with at least one of our methods, or remain extremely competitive, to within at most 5 F1 points---the difference is usually within fractions of a point.

In coreference tasks, choice of metric reflects heavily in the results. For instance, almost 33\% of the ECB+ dataset across all three splits consists of singleton mentions, and MUC score is not as sensitive to the presence of singletons as $B^3$. On the other hand, $CEAF_e$’s alignment algorithm tends to ignore correct coreference decisions when response entities are misaligned~\cite{moosavi2016coreference}.

\section{Zero-Shot Prompt Design}
\label{app:prompt}

The prompt formats used for zero-shot evaluation of LLaMA 2-7B-Chat and GPT-3.5-Turbo are given below.

\begin{tcolorbox}[title=\textsc{LLaMA 2-7B-Chat Zero-shot Prompt Format}]
\begin{dialogue}
	\speak{\sc system\_prompt:} Think step by step. You are a coreference annotator and you have to make a decision about two events marked with {\tt <m>} and {\tt </m>} tokens. You are given two sentences. Answer in one word if they are talking about different events or the same event.
			 
	\speak{\sc user\_prompt:} 	Sentence 1 is: \{sentence\_1\}. Sentence 2 is: \{sentence\_2\}. Your answer: 
\end{dialogue}
\end{tcolorbox}

\begin{tcolorbox}[title=\textsc{GPT-3.5-Turbo Zero-shot Prompt Format}]
\begin{dialogue}
	\speak{\sc system\_prompt:} You are a coreference annotator and you have to make a decision about two events marked by {\tt <m> }and {\tt </m>} tokens. You are given two sentences. Answer in one word if they are talking about the same event: that is, if they are coreferent.
			 
	\speak{\sc user\_prompt:} sentence\_1: \{sentence\_1\}\\ sentence\_2: \{sentence\_2\} 
\end{dialogue}
\end{tcolorbox}

\section{Quality of Free Text Rationales}
\label{app:im-quality}

To evaluate the quality of information presented through the rationales, we presented human evaluators with a set of questions about a small sample of generated FTRs. The questions were inspired by \citet{wiegreffe2021measuring} and sought to assess factors like fact content, relevancy, plausibility, and the quality of the reasoning process demonstrated in the written output, according to humans.  

\begin{figure}[h!]
    \centering
    \includegraphics[width=.5\textwidth]{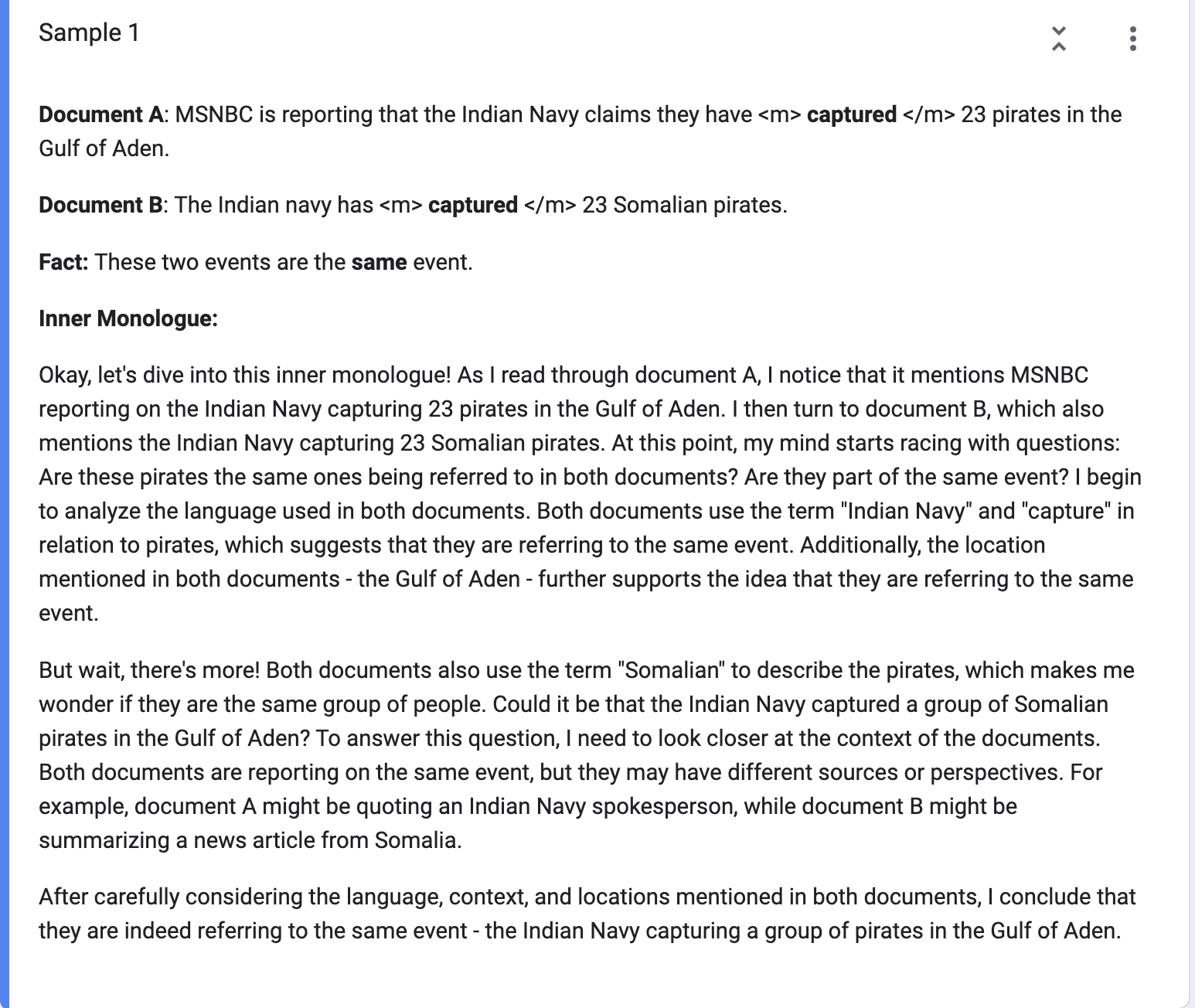}
    \caption{Rationale sample presented to evaluators.}
    \label{fig:rationale-eval}
\end{figure}

Four evaluators (all adult English speakers) took a survey containing pairs of event mentions from two different documents (6 pairs each drawn from the ECB+ and GVC corpora), the ground truth label (which was also given to LLaMA 2-7B-Chat for generation), and the generated inner monologue FTR (see Fig.~\ref{fig:rationale-eval}). They were asked to answer seven multiple choice questions for each sample, designed to explore various aspects of the inner monologue-based explanation.

The questions included:

\begin{itemize}
    \item {\bf Factuality}: How factual is this Inner Monologue-based explanation? ({\it generally true}/{\it partially true}/{\it generally false}/{\it not enough information})
    \item {\bf Relevance}: Is the Inner Monologue-based explanation relevant to the context? ({\it yes}/{\it no})
    \item {\bf New Information}: Does the Inner Monologue sample provide new facts, information, or reasoning not stated in the pair of documents explicitly? ({\it yes}/{\it no})
    \item {\bf New Information Relevance}: If you answered yes to the above question, is the new information or reasoning relevant to justifying the facts about the events? ({\it yes}/{\it no}/{\it not enough information})
    \item {\bf Gold Label}: How much information does the Inner Monologue sample have to justify the facts about the two events? ({\it enough}/{\it not enough}/{\it more than enough}/{\it can't say})
    \item {\bf Plausibility}: Is the Inner Monologue sample acceptable or plausible considering the context? ({\it yes}/{\it no}/{\it can't say})
    \item {\bf Inner Speech Overlap}: If you were to use your own inner-monologue-based reasoning to arrive at the FACT, how much of an overlap does your thought-pattern have with the given Inner Monologue? ({\it high overlap}/{\it some overlap}/{\it minimal overlap}/{\it no overlap})
\end{itemize}

Annotations were performed by members of the authors' research lab in the course of their normal lab duties. A different set of 4 annotators were used to assess samples from each corpus---2 male and 2 female in each set, drawn from a university population. Annotators had some casual exposure to the problem of CDCR but no other prior experience in the task. The survey was determined to be Not Human Subjects Research by the institutional review board.

\begin{figure}
    \centering
    \includegraphics[width=.5\textwidth]{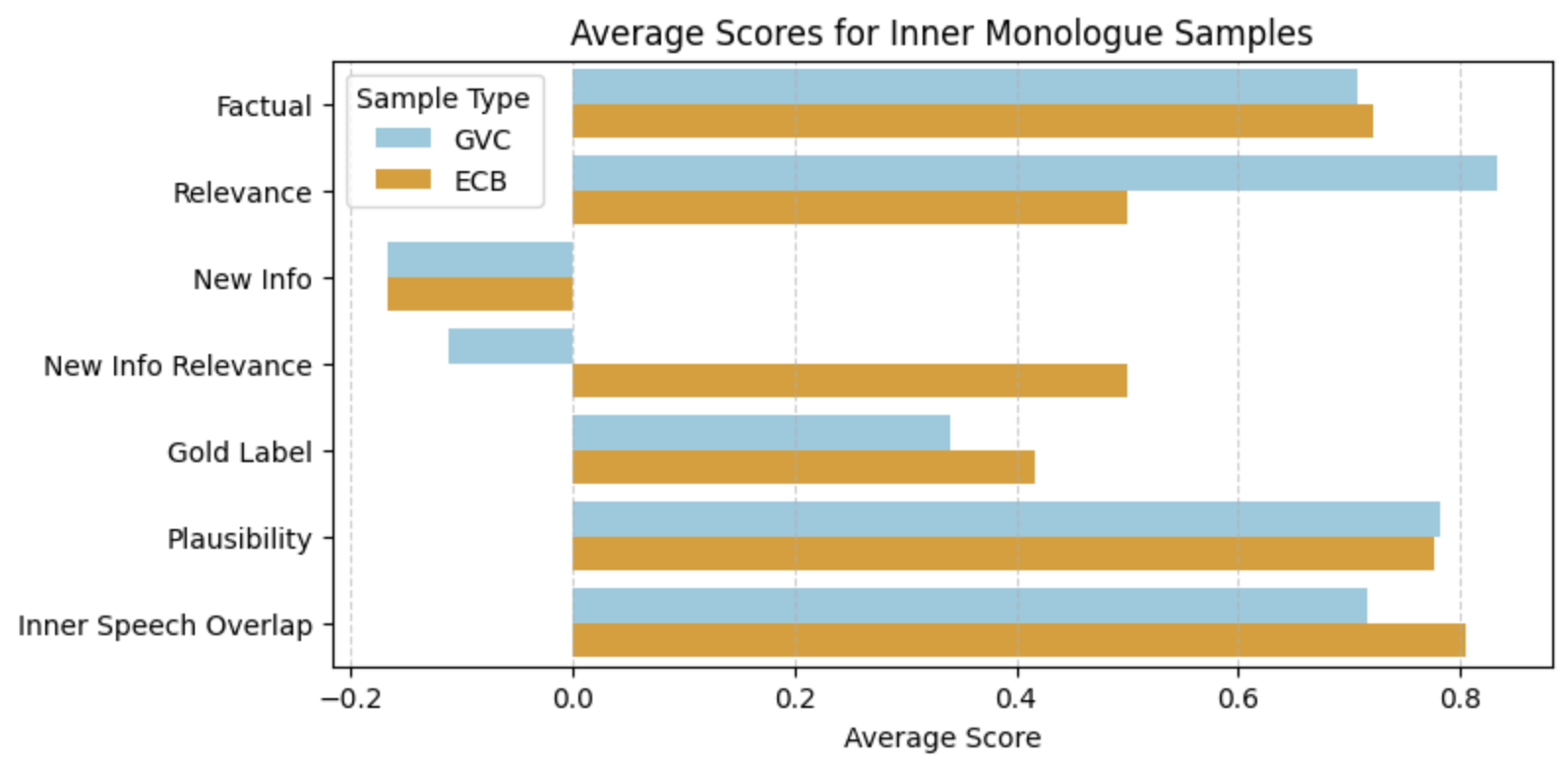}
    \caption{Average scores for inner monologue samples generated from ECB+ and GVC.}
    \label{fig:gvc-ecb-eval}
\end{figure}

The answers were then mapped to numerical values following the template of \citet{wiegreffe-etal-2022-reframing}.  Yes/no answers were mapped to -1/1.  Answers to the multiple-choice questions were mapped to -1 (negative valence), 0 (uninformative/neutral), 0.5 (partially positive valence, where available), or 1 (positive valence). Figure~\ref{fig:gvc-ecb-eval} shows average scores for inner monologue samples on the above questions.

FTRs generated from both datasets were rated as highly factual, showing that they were representing accurate information about the events concerned. GVC FTRs were rated as more highly relevant than ECB+ FTRs. This may reflect the low topic diversity of GVC (since all events concern gun violence, as long as the FTR remains on-topic, it remains relevant). Negative scores for new information on both datasets indicate a challenge in generating content not already mentioned in the source material. This suggests a potential area for improvement in terms of content generation fidelity. Where the FTR introduced new information, evaluators of ECB+ FTRs found this information more relevant than evaluators of GVC FTRs (this may also be a sparsity effect). FTRs from both datasets were rated as highly plausible, indicating similar levels of logical coherence in their inner monologue samples. When asked to assess the level of overlap between how the FTR proceeded and how they would think about the question if using inner speech, evaluators rated this highly.

We calculated Krippendorff's $\alpha$ \cite{krippendorff2011computing} as a metric of evaluator agreement.  We found that $\alpha \approx .22$ for ECB+ FTRs and $\alpha \approx .06$ for GVC FTRs. While the average scores above indicate that the generated FTRs appear to contain information for coreference decisions that humans consider useful, the annotator agreement scores indicate the subjective nature of the evaluation task. Conditioning generation on the gold label, and providing the label to the annotators, places controls on human evaluation of rationales, since annotators tends to inject bias against rationales when they disagree with the gold label~\cite{wiegreffe-etal-2022-reframing}. Untrained humans often weigh the same information significantly differently in the same task \cite{zhao-etal-2023-cross}.

\section{Okay, Let's Do This?}

The title of our paper comes from the fact that LLaMA 2-7B-Chat begins its FTRs with a "chatty" introductory sentence before starting to generate content regarding the event pair.  "Okay, let's do this!" is one of the most frequent (and amusing) introductory sentences occurring in our FTR corpus.
\end{document}